\documentclass{article}

\usepackage[final, nonatbib]{NFFL2021}

\usepackage[utf8]{inputenc} 
\usepackage[T1]{fontenc}    
\usepackage{hyperref}       
\usepackage{url}            
\usepackage{booktabs}       
\usepackage{amsfonts}       
\usepackage{nicefrac}       
\usepackage{microtype}      
\usepackage{xcolor}         
\usepackage{wrapfig}
\usepackage{amsmath}
\usepackage{caption}
\usepackage{subcaption}
\usepackage{graphicx}
\usepackage{enumitem}
\usepackage{tikz}
\usepackage{varwidth}

\title{Certified Federated Adversarial Training}

\author{%
  Giulio Zizzo \\
  IBM Research Europe \\
  \texttt{giulio.zizzo2@ibm.com} \\
  \And
  Ambrish Rawat \\
  IBM Research Europe\\
  \texttt{ambrish.rawat@ie.ibm.com} \\
  \And
  Mathieu Sinn \\
  IBM Research Europe\\
  \texttt{mathsinn@ie.ibm.com} \\
  \And
  Sergio Maffeis \\
  Imperial College London\\
  \texttt{sergio.maffeis@imperial.ac.uk} \\
  \And
  Chris Hankin \\
  Imperial College London\\
  \texttt{c.hankin@imperial.ac.uk} \\
}

\begin{document}

\maketitle

\begin{abstract}
In federated learning (FL), robust aggregation schemes have been developed to protect against malicious clients. Many robust aggregation schemes rely on certain numbers of benign clients being present in a quorum of workers. This can be hard to guarantee when clients can join at will, or join based on factors such as idle system status, and connected to power and WiFi~\cite{kairouz2019advances}. We tackle the scenario of securing FL systems conducting adversarial training when a quorum of workers could be completely malicious. We model an attacker who poisons the model to insert a weakness into the adversarial training such that the model displays \emph{apparent} adversarial robustness, while the attacker can exploit the inserted weakness to bypass the adversarial training and force the model to misclassify adversarial examples. We use abstract interpretation techniques to detect such stealthy attacks and block the corrupted model updates. We show that this defence can preserve adversarial robustness even against an adaptive attacker.
\end{abstract}

\section{Introduction}

Training via federated learning (FL)~\cite{mcmahan2017communication} is increasingly popular due to the many strengths of FL, which include reducing communication overheads, decentralising computations, and preserving data privacy. However, FL offers new attack opportunities to an adversary. Attackers can participate in FL and pursue objectives such as inserting backdoors~\cite{xie2020dba} or preventing model convergence~\cite{baruch2019alie}. To combat this, robust aggregation schemes have been developed~\cite{blanchard20017machine, yin2018byzantine, xie2019zeno, xie2018generalized}. These algorithms offer protection if the proportion of malicious clients does not exceed particular thresholds. For example, median based robust aggregation schemes require the majority of the clients participating in a round of FL to be benign~\cite{yin2018byzantine}. However, having this assurance in practical settings at all times can be difficult to maintain: benchmark FL systems~\cite{caldas2018leaf} have many thousands of users, of which only a small proportion participate in any one FL round. Therefore, if clients are modelled as randomly participating, or worse if they can join at will, then it is likely that a round of FL will occur in which the quorum of participating clients is maliciously dominated breaking robust aggregation defences.

In parallel, it is known that neural networks are vulnerable to adversarial examples, which in turn can be mitigated using adversarial training \cite{madry2017towards}. The interaction of adversarial training with FL is an active area of research with results showing federated adversarial training's sensitivity to the amount of local compute \cite{shah2021adversarial}, that not all clients need to necessarily perform adversarial training to achieve robustness \cite{hong2021federated}, as well as specialised attacks against federated adversarial training \cite{zizzo2020fat}.

In this work we consider how federated adversarial training could be conducted if malicious clients can dominate particular quorums. Specifically, the scenario we consider is:

\begin{itemize}[leftmargin=*]
    \item A defender conducts FL in a large distributed setting, and only a small number of clients participate in any one FL round. The defender cannot guarantee that a particular quorum of workers is not maliciously dominated.
    \item The defender seeks a robust model and so instructs the clients to conduct adversarial training.
\end{itemize}

We consider two research questions arising from this scenario: how can a defender protect themselves against stealthy compromise of their model? Additionally, to what extent can an adaptive attacker overcome a proposed defense?
The contributions we make in addressing the above questions are: 
\begin{itemize}[leftmargin=*]
    \item We demonstrate the effectiveness of certification based on abstract interpretation techniques as a defensive method in federated learning. Due to the guarantees that abstract interpretation can provide, stealthy model tampering can be instantly detected.
    \item We propose a novel attack strategy that can be employed by an adaptive attacker to try and evade certification defences. The attack exploits the potentially limited number of datapoints a defender can verify. So, the tampered model that the attacker sends \text{only} certifies the datapoints that the defender checks, while still being stealthily weak on all other data.
\end{itemize}

\section{Background}

\subsection{Federated Learning}

In FL many clients wish to learn a common model, and each client trains using their local data. Once the clients have updated their local models they send their trained weights to an aggregator. The aggregator combines the received weights producing an updated model. The model is then sent to the clients for a new training round. However, malicious clients can propose arbitrary updates pursuing objectives such as preventing convergence~\cite{baruch2019alie} or inserting backdoors~\cite{xie2020dba}. To that end robust aggregation schemes have been developed to filter out malicious updates~\cite{blanchard20017machine, yin2018byzantine, xie2019zeno, xie2018generalized}.

Adversarial training for federated learning is vulnerable to convergence attacks~\cite{zizzo2020fat}. However, we focus on a more dangerous kind of attack~\cite{zizzo2020fat}, where an adversary replaces a model which is undergoing adversarial training with one that is only superficially robust, for example due to gradient masking. In this case, the defender can be unaware that the model has been compromised.

Robust aggregation defences rely on bounding the number of malicious clients $k$ present in a quorum of $n$ clients. Median based defences handle at at most $\lceil  \frac{n}{2} \rceil - 1$ malicious clients~\cite{yin2018byzantine}, Krum needs $n > 2k+2$ \cite{blanchard20017machine}, or Bulyan \cite{guerraoui2018hidden} requires $n \geq 4k+3$. However, benchmark datasets for simulating real FL have a very small proportion of clients participating in a round~\cite{caldas2018leaf}. This raises difficulties for such defences as, even if there is a small proportion of adversaries in a system, it is likely that given sufficient training rounds, there will be rounds in which the threshold requirements for robust aggregation schemes are broken. Therefore, to offer protection in this challenging environment, we turn to certification based approaches.

\subsection{Certification}

Certification of neural networks is a research area orthogonal to FL. Via certifiable methods a defender has a \emph{guarantee} whether a datapoint can have its prediction changed, which can be used to uncover stealthy model compromise. More precisely, given a neural network $f$ the defender wishes to verify that the predicted label for all inputs from a $p$-norm ball $B^p_\epsilon$ of size $\epsilon$ centred on the datapoint $x$ are classified to the same class $T$,

\begin{equation}
\label{equ:certification}
\text{arg max} f(x^\prime) = T \qquad    \forall x^\prime \in B^p_\epsilon(x) = \{x^\prime = x + \delta \: | \; \; |\delta|_p \leq |\epsilon|_p \}.
\end{equation}

If such a condition holds then a datapoint is certified. We also refer to certifying neural networks by measuring the proportion of datapoints it can certifiably classify. There are several research strands which try and verify adversarial robustness through certification \cite{cohen2019certified, tjeng2017evaluating}. In this work we model a defender using abstract interpretation methods to verify the adversarial robustness of the model. With abstract interpretation we pass through the neural network a datapoint's abstract representation in a domain such as interval \cite{gehr2018ai2}, zonotope \cite{singh2018fast}, octagon \cite{mine2006octagon}, or polyhedra \cite{singh2019abstract}. More precisely, we can represent a concrete input $x$ with all its possible perturbations with an abstract element $\hat{x}$. The abstract element will capture the entirety of $B^p_\epsilon$. By passing $\hat{x}$ through the neural network, and analysing its abstract output $\hat{y}$, we can check if a robustness property holds for the network.

The various abstract representations often trade precision for scalability. The zonotope domain was shown to offer strong performance while remaining computationally tractable for neural networks of realistic size, and is the domain we will focus on \cite{singh2018fast}. A zonotope $\hat{x}$ is represented by 

\begin{equation}
    \hat{x} = \eta_0 + \sum_{i=1}^{i=N} \eta_i \epsilon_i 
\end{equation}

where $\eta_0$ corresponds to a central coefficient and $\eta_i$ represents $i=1 \dots N$ deviations around the center with associated error symbols $\epsilon_i$. The values $\epsilon_i$ can take either +1 or -1. Pushing zonotopes through neural networks requires computing \textit{abstract transformers} for various arithmetic operations. Many operations, like matrix multiplication, can be carried to the abstract domain easily. However, certain operations require more consideration. For the neural networks we consider, ReLUs can cause the analysis to become more imprecise. To balance scalability and precision in our abstract transformer for ReLUs we use the DeepZono formulation proposed in~\cite{singh2018fast}. To help build intuition for the reader into using zonotopes for neural networks we show a toy example in Appendix A.

\textbf{Certification in FL:} There have begun to be several works which examine certifying a FL trained model against backdoors. Parallel work to ours \cite{xie2021crfl} examined parameter smoothing to certify robustness against $L_2$ backdoors. General provable robustness was investigated in \cite{cao2021provably}, which although protects against \emph{any} type of adversary, it required training hundreds of parallel models with tight requirements on the number of malicious clients. In contrast, centralised training was examined in~\cite{weber2020rab} for certification against backdoors via training several models on noise corrupted data. This poses problems in FL as malicious clients may not conduct the prescribed training procedure.

\textbf{Certified Training:} It is worth considering the case in FL where the clients perform certified training \cite{mirman2018differentiable, balunovic2019adversarial} rather than normal adversarial training. We assume the defender does not pursue this for a few reasons. Firstly, performing certified training causes a drop in accuracy on normal data: this is on top of the drop in accuracy that occurs with projected gradient descent (PGD) training~\cite{balunovic2019adversarial}. Additionally, certified training methods can result in worse white-box PGD performance \cite{mirman2018differentiable}. Recent work~\cite{balunovic2019adversarial} is bridging this gap, but conducting certified training in FL is an open challenge. Secondly, certified training itself has many open questions which should be addressed prior to making it federated such as the conflict between more accurate abstractions leading to worse certifiable results~\cite{balunovic2019adversarial}.

Having established the background theory we now specify our attacker model more precisely, detail our proposed defence, and finally discuss adaptive attacks against it.

\section{Attacker Model}
\label{sec:attacker_model}
\textbf{Attacker Objective:} We consider an attacker who wishes to stealthily compromise the adversarial training procedure. Specifically, they wish their compromised model to have the following properties: 
\begin{itemize}[leftmargin=*]
    \item High \emph{normal} data performance.
    \item High adversarial robustness when the model is evaluated with standard application of PGD to the maximum allowable $L_\infty$ budget that the defender will evaluate to.
    \item A hidden weakness that the attacker can exploit such that they can drive the adversarial performance of the model down when still limited to modifying the data within the given $L_\infty$ budget.
\end{itemize}

There are several ways the above could be achieved. Firstly, a backdoor with a $L_\infty$ bounded trigger could be inserted. The attacker will train the model adversarially, but also insert the backdoor. When the model is presented with the backdoor trigger, even if it was adversarially trained, the model will misclassify to an attacker chosen class. Backdoor examples, with discussion on $L_\infty$ backdoors, are shown in Appendix B. Alternatively, a stronger style of attack is to pursue gradient masking based strategies such as sending a defensively distilled model to the aggregator~\cite{zizzo2020fat}. The defender, unaware that they have a model with defensive distillation will see, through standard application of PGD, that the model is apparently robust. An attacker, who deliberately sent the defensively distilled model, can then bypass the brittle defence by application of temperature scaling to the softmax~\cite{carlini2016defensive}. As the distilled model does not have a backdoor trigger it can sidestep many methods which rely on backdoor detection and certification~\cite{xie2021crfl}. A review of defensive distillation and its use as an attack method is outlined in Appendix C.

\textbf{Attacker Knowledge and Capabilities:} The malicious clients cannot determine when they participate in a FL round, but when participating they can co-ordinate between each other. For our basic adversary we assume the attacker has access to all the data held by the malicious clients, knows the robust aggregation rules, and knows the benign model updates that are sent to the aggregator.

\section{Certified Defence}
\label{sec:cert_defence}
The defender wishes to conduct federated adversarial training and, being aware of potential attackers, does not blindly accept model updates and will employ a robust aggregation algorithm. In addition, the defender has access to a held out set of data (similar defender modelling assumptions were used in \cite{xie2019zeno}) and rejects models which perform poorly on it, both in terms of normal and adversarial accuracy against standard white box PGD attack \cite{madry2017towards}. The new component we propose is a certification based assertion to detect stealthy compromise. More precisely, the defender checks:
\begin{enumerate}[leftmargin=*]
    \item The certified accuracy over a certification dataset which is drawn from a held out set of data. We define the certified accuracy as the proportion of data that satisfies the certification condition given by Equation \ref{equ:certification}, and in addition, is also classified correctly by the neural network. 
    \item Secondly, as we will later see, the certifiable accuracy is often not sensitive enough to distinguish between a corrupted vs adversarially trained model. As a more sensitive metric the defender will also use the certifiable loss. We use a modified version of the loss in~\cite{mirman2018differentiable} of, 
    \begin{equation}
    \label{equ:cert_loss}
    L(z, y) = \text{max}(0, \underset{q \neq y}{\text{max}}(z_q - z_y))
\end{equation}

where $z_q$ is the component of the zonotope associated with the logit for label $q$ and $y$ is the target label. Computing the max operation over all the $z_q - z_y$ pairs can be computationally unfeasible as each zonotope component can have thousands of $\epsilon$ error terms. Hence, we approximate and compute the max operation over the interval representation of $z_q - z_y$.
\end{enumerate}

Each round the defender will collect the relevant metrics (normal accuracy, adversarial accuracy, and mean certifiable loss and certified accuracy) and will accept the updates if they have at least 90\% of the various accuracy scores from the prior round, and  the certifiable loss has varied by a maximum of 10\%. This is with the assumptions that attackers can only join randomly and so not coordinate over multiple rounds to slowly drive down model performance. More sophisticated filtering strategies can clearly build on this baseline.

If the aggregated model during a FL round passes all of these criteria then it will be accepted by the aggregator, else the round is skipped.

\section{Adaptive Attacks Against Certification}
\label{sec:stealth_attack}

We propose a novel adaptive attack which can be employed by a strong adversary who has knowledge of the certification defence to try and evade it. The certification algorithm itself is unassailable: by definition of the soundness of the abstract interpretation methods, it is impossible to have the certification algorithm claim a datapoint cannot be misclassified and then, through any attack algorithm, misclassify that datapoint. Hence, the attacker will need to consider alternative attack vectors. 

The attack takes advantage of the fact that certification approaches can be slow and computationally expensive. The DeepZono transformation which we use is a \emph{fast} certification, and yet can take a few seconds for complex data: in the original paper which proposed it \cite{singh2018fast} only the first 100 samples of the MNIST and CIFAR10 test datasets were used for the results. Hence, the attacker can exploit the limited number of samples that could be realistically checked by a defender. Still maintaining the original three attack objectives specified in Section \ref{sec:attacker_model}, we add an additional goal of:
\begin{itemize}
    \item Match the certification metrics over the datapoints that the defender will check.
\end{itemize}

The adaptive attacker knows the exact certification set, in addition to any other data that the defender uses to check the model performance. This represents the strongest possible adversary with perfect knowledge. By evaluating our defence under this threat model we can obtain measures of its worst case performance. The attacker pursues their goal via the following methodology:
\begin{enumerate}[leftmargin=*]
    \item Train a model with defensive distillation. Such a model will fulfil the original goals defined in Section \ref{sec:attacker_model}, but will be detected by the certifiable check as being vulnerable.
    \item Refine the defensively distilled model obtained from Step 1. The attacker optimises for the model to match the certification statistics that a defender uses to detect stealthy model compromise: either certified accuracy or, in addition, the mean certifiable loss. However, just optimising for this objective removes the effect of defensive distillation. Therefore, the attacker jointly optimises to match the certifiable performance and maintain defensive distillation. The loss function to achieve both certifiable accuracy and match the mean certifiable loss is given by Equation \ref{equ:cert_loss}.
\end{enumerate}

Based on our experiments when developing this attack, this strategy does not work if the adversary pursued a backdoor based attack to bypass the adversarial training: the backdoor would be destroyed and a robust model would remain. We further experimented with instead conducting PGD training over the certification set in Step 2 to try and achieve the same certification metrics. However, this was unsuccessful. Although the model would classify the certification set correctly the certified accuracy and loss remained poor, and so the model would be detected by a defender as being compromised.

\section{Experiments}

We use the MNIST and CIFAR10 datasets. For both datasets we use 500 clients, each of which is randomly assigned a dataset partition. We set 175 clients to be malicious (35\% of the total number). The neural networks used for each dataset are shown in Appendix D. In each round of training 5 clients are randomly selected to participate. The aggregator uses median based aggregation tolerating up to 2 malicious clients.  We will be dealing with $L_\infty$ bounds in two different contexts: either the bound used for certification or the bound used to craft adversarial examples. To aid in disambiguation in the notation we use $L_\infty^{crt}$ for the bound to which certification is conducted, and $L_\infty^{adv}$ is the bound to which adversarial examples are constructed. For MNIST the models train adversarially with $L_\infty^{adv} = 0.25$, and for CIFAR10 we use $L_\infty^{adv} = 8 / 255$. The defender uses the first 1000 datapoints of the MNIST training dataset and the first 100 datapoints of the CIFAR10 training dataset as the certification datasets to certify the aggregated models. This is to have similar compute times spent on the certification defence with both datasets. Then, to compute the normal and adversarial accuracy scores in addition to the certification statistics each round (as described in Section \ref{sec:cert_defence}) the defender has a validation dataset. For this, the defender uses the next 5000 points from the MNIST/CIFAR10 training sets. Neither the validation nor certification data is used for client training. 

The attacker waits until the model has neared convergence and at least 3 malicious clients are selected to participate in a FL round. At that point the adversary has control of the system and runs the attack. The attacker runs either the distillation or backdoor attacks described in Section \ref{sec:attacker_model} to stealthily weaken the adversarial training. To match the PGD performance we use a temperature scaling for the defensive distillation of $T=100$ for MNIST and $T=25$ for CIFAR10. The backdoor trigger is composed of vertical stripes of magnitude $L_\infty = 0.1$ for MNIST and $6/255$ for CIFAR10 keeping the triggers below the adversarial training limit.

\subsection{Certification in Federated Learning}

We first demonstrate that against an non-adaptive attacker, a defender which cannot assure that the benign client thresholds for robust aggregation schemes are met can use certification approaches to detect stealthy model compromise. The results for MNIST are shown in Table~\ref{tab:mnist_results}. Our first take-away is that relying solely on the certifiable accuracy as a metric gives a crude from of defence: although we adversarially trained with PGD at $L_\infty^{adv} = 0.25$ we do not have any certifiable accuracy at that level, and so leaves us blind to the differences between an honestly trained PGD model and the two compromised models. However, we \emph{can} use the mean certifiable loss (Equation \ref{equ:cert_loss}) over our certification points as a metric to distinguish between the models at $L_\infty^{adv} = 0.25$. At lower certification bounds (eg $L_\infty^{crt}=0.1$ and $0.15$) a adversarially PGD trained model can have good certifiable accuracy. Hence, at those lower certification bounds both mean certifiable loss and accuracy differ significantly for an adversarially PGD trained model and the two attacker manipulated models.

For CIFAR10 we have a similar pattern: the certifiable accuracy is a good metric to distinguish between models at low certification bounds of $L_\infty^{crt}=1/255$. When using the mean certifiable loss there is a large difference at all bounds making it simple to identify the tampered models.

\begin{table}[]
    \centering
    \setlength\tabcolsep{5pt}
    \begin{tabular}{c@{\hspace{4pt}} c c@{\hspace{4pt}} c c c c c c c}
        \toprule
        Dataset & Model & \multicolumn{2}{c}{Acc} & \multicolumn{6}{c}{Certified}  \\
                &       &   Normal       &  Adv       &          Acc & Loss             &              Acc & Loss          & Acc & Loss \\\cline{5-10}
                \addlinespace[0.1cm]
                &       &          &  $L_\infty^{adv} = 0.25$  & \multicolumn{2}{c}{$L_\infty^{crt} = 0.1$} & \multicolumn{2}{c}{$L_\infty^{crt} = 0.15$} &  \multicolumn{2}{c}{$L_\infty^{crt} = 0.25$} \\
                \midrule
                &    PGD        &  98.78   &  92.03  & 91.4 &  0.15  & 33.4  & 2.72 &  0.0 & 32.96 \\
        MNIST   &    Distilled  & 97.23 & 95.22  & 0.0  & 8.5x$10^3$  &  0.0 &  2.0x$10^4$  &  0.0 & 4.4x10$^4$ \\ 
        &    Backdoor   & 98.81 & 94.10  & 0.0 & 50.81 & 0.0 & 96.47  &  0.0 & 202.43 \\\cline{5-10}
                \addlinespace[0.1cm]
                &      &             &  $L_\infty^{adv} = 8/255$      & \multicolumn{2}{c}{$L_\infty^{crt}=1/255$} & \multicolumn{2}{c}{$L_\infty^{crt} = 4/255$} &  \multicolumn{2}{c}{$L_\infty^{crt}=8/255$} \\
                \addlinespace[0.02cm]
                \cline{5-10}
                \addlinespace[0.1cm]

                &    PGD       & 72.71 & 40.24 & 60.0 & 0.13 & 0.0 & 596.0 & 0.0  & 5.4x10$^3$  \\
         CIFAR &    Distilled  & 78.35 & 44.17 & 0.0 &  2.1x10$^6$ & 0.0 & 9.5x$10^7$ & 0.0 & 3.7x10$^8$  \\
               &    Backdoor   & 74.70 & 32.47 & 41.0 & 1.14 & 0.0 & 2.3x10$^3$ & 0.0 & 1.5x10$^4$ \\
    \bottomrule
    \addlinespace[0.1cm]
    \end{tabular}
    \caption{At lower certification bounds large differences between a model trained via PGD compared to the distilled and backdoored models are visible in terms of both the certified accuracy and the mean loss across the certification set. At higher bounds only the mean certifiable loss is a viable metric.}
    \label{tab:mnist_results}
\end{table}

\subsection{Adaptive Attacker}

We now examine an adaptive attacker who creates a brittle robustness model which certifies the defender's certification dataset. We use the strongest possible attacker model defined in Section \ref{sec:stealth_attack} to push the defence as far as we can to examine its worst case performance. As discussed in Section \ref{sec:stealth_attack}, the backdoor strategy is ineffective for an adaptive attack so an adversary uses the distillation attack and jointly optimises to match the certification statistics on the certification data. We show adaptive attack results for MNIST. For CIFAR10 the adaptive attack was unsuccessful: the classification task is more challenging, and optimising with the adaptive objective did not give a functioning model. Larger models may have the capacity for this objective. Such investigations are left for future work.

\subsubsection{Simpler Defender: Only Certifiable Accuracy Considered}

\begin{figure}
    \centering
    \includegraphics[trim={2cm 0 2cm 0}, clip, width=0.95\textwidth]{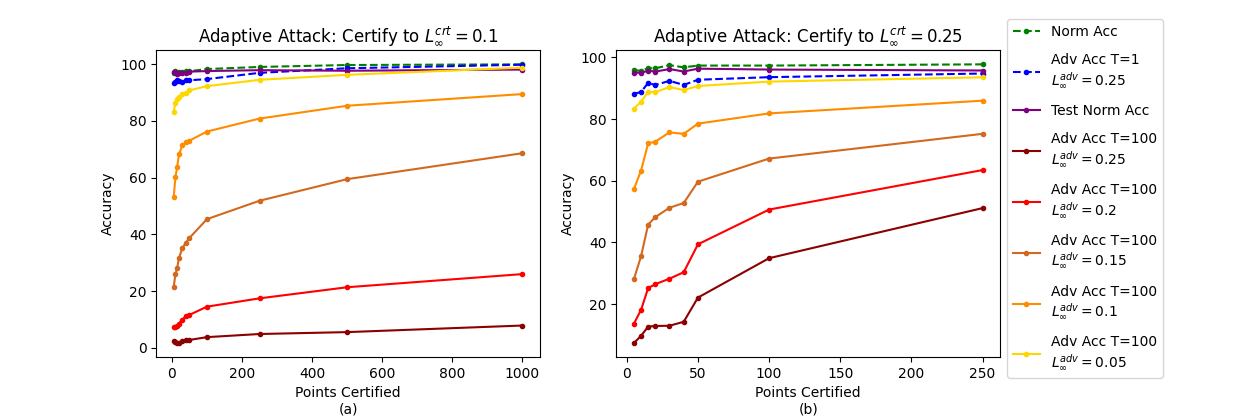}
    \caption{Performance of the adaptive attack. The green line shows the malicious model's performance on the clean validation set and the blue line represents what the defender sees when evaluating it with PGD on their validation set. The purple line is the accuracy of the malicious model on the clean MNIST test set. The yellow through to dark red hues represent the actual robustness of the model with differing $L_\infty^{adv}$ budgets on the MNIST test set. On the $x$ axis we have the number of datapoints from the defender's certification set that the attacker tries to either 1) certifiably classify in the left plot or 2) match the mean certifiable loss over in the right plot.}
    \label{fig:cert_results}
\end{figure}

We first consider a defender who just checks the certified accuracy of a model to demonstrate that this is only effective against an adaptive attacker at low $L_\infty^{adv}$ bounds. In Figure~\ref{fig:cert_results}a the defender checks the certifiable accuracy at $L_\infty^{crt} = 0.1$, as at higher bounds such as $L_\infty^{crt} = 0.25$ the certifiable accuracy is not a distinguishing metric between a genuinely trained PGD model or a tampered model. We show the resulting performance when the adaptive attacker certifiably classifies varying amounts of the defender's certification dataset to $L_\infty^{crt} = 0.1$. The resulting model is evaluated with PGD white box attack with budgets ranging between $L_\infty^{adv} = 0.05 - 0.25$.

Each graph shows 1) the accuracy on normal clean data, 2) the accuracy on adversarial data with a temperature $T=1$ on the softmax (no temperature effects) that the defender sees when evaluating with PGD, and 3) the performance of the model when attacked by an adversary using temperature scaling with $T=100$ to bypass the defensive distillation. The adversarial examples crafted when using $T=100$ are then evaluated on a model with $T=1$ that a target client would be using.

We can see how the performance of the adaptive attack varies with the $L_\infty^{adv}$ budget for crafting adversarial examples. If the $L_\infty^{adv}$ budget is large in relation to the certified bound then the defence is ineffective. We see in Figure \ref{fig:cert_results}a the defender will see a model that fulfils their criteria of 1) high normal accuracy (green points), 2) high white box PGD performance (blue points) and 3) has the correct certification metrics. By applying temperature scaling when making an adversarial example, at high crafting budgets (eg dark red points), the model has poor actual adversarial robustness. 

However, if the defender accepts lower levels of adversarial robustness then the adaptive attack boosts the underlying robustness of the model. As more samples are forced to be certifiably correct, then the adversarial robustness improves and generalises across the test set. Therefore, by considering certifiable accuracy, robustness to $L_\infty^{adv}=0.25$ cannot be maintained against a adaptive attacker. However, this method does ensure robustness at lower bounds (e.g. $L_\infty^{adv}= 0.1$) \emph{can} be achieved.

\subsubsection{Full Defender: Certifiable Accuracy and Loss Considered}

We now consider the more sensitive approach of using the certifiable loss which gives a signal at higher certification bounds to detect model compromise. We examine the performance when conducting certification on $L_\infty^{crt} = 0.25$ in Figure \ref{fig:cert_results}b. As a PGD trained model is unable to certifiably classify any of the certification set, the defender considers the mean certifiable loss which the attacker attempts to match. With a bound of $L_\infty^{crt} = 0.25$ conducting the adaptive attack is more challenging and immediately pursuing this with $L_\infty^{crt} = 0.25$ was ineffective. Instead we first certify the datapoints to a bound of 0.1 and grow the certification bound in steps of 0.01 when the attack succeeds in matching the mean certifiable loss.

The results are shown in Figure \ref{fig:cert_results}b. The adaptive attack was unable to proceed to larger certification sets: at 500 data points the attack stalled and did not find a solution after 72 hours of optimising. Further, the more datapoints the defender uses results in the neural network becoming \emph{genuinely} robust if the attacker optimisation succeeds. By certifying an increasingly larger dataset the attacker is forced to impart real adversarial robustness into the model. Therefore, empirically the defence was robust with 1000 certification points for MNIST when subject to the strongest attacker we developed. Further, even if the attacker had unlimited computing speed and could certify larger datasets, from the trends in Figure \ref{fig:cert_results}b, the model's robustness will continue rising making the attacker less effective.

In addition to stalling at 500 certification points, the attack failed against CIFAR10. We postulate that due to the complexity of the attack balancing two contrasting objectives, the model capacity can be insufficient to solve the objectives. Therefore, the defender can "out compute" an adaptive attacker by certifying a number of datapoints for which an attacker is unable to optimise for. Future work would be to determine, for a given model, what the minimum certification set size is, such that even if the attacker optimisation succeeds, it results in a genuinely robust model. This is because the defender wishes to minimise the defence compute cost while ensuring that their model is safe.

\section{Conclusion}

We have examined the problem in FL of maliciously controlled quorums. There are many security problems that arise from this, and we consider the case of securing adversarial training. We have shown that certified methods form an effective defence. Adaptive attackers can force a manipulated model to have certified performance on a certification dataset and have hidden adversarial example weakness. However, this attack has a high compute cost making it impractical for more then a limited number of certification points. Additionally, with increasing certification data the attack strengthens the model's underlying adversarial robustness.

\section*{Acknowledgments}

This project has received funding from the European Union’s Horizon 2020 research and innovation
programme under grant agreement No 824988. https://musketeer.eu/. 

\bibliographystyle{plain}
\bibliography{refs.bib}

\begin{thebibliography}{10}

\bibitem{balunovic2019adversarial}
Mislav Balunovic and Martin Vechev.
\newblock Adversarial training and provable defenses: Bridging the gap.
\newblock In {\em International Conference on Learning Representations}, 2019.

\bibitem{baruch2019alie}
Gilad Baruch, Moran Baruch, and Yoav Goldberg.
\newblock A little is enough: Circumventing defenses for distributed learning.
\newblock In {\em Advances in Neural Information Processing Systems 32: Annual
  Conference on Neural Information Processing Systems 2019, NeurIPS 2019,
  December 8-14, 2019, Vancouver, BC, Canada}, pages 8632--8642, 2019.

\bibitem{blanchard20017machine}
Peva Blanchard, El~Mahdi~El Mhamdi, Rachid Guerraoui, and Julien Stainer.
\newblock Machine learning with adversaries: Byzantine tolerant gradient
  descent.
\newblock In {\em Advances in Neural Information Processing Systems 30: Annual
  Conference on Neural Information Processing Systems 2017, December 4-9, 2017,
  Long Beach, CA, {USA}}, pages 119--129, 2017.

\bibitem{caldas2018leaf}
Sebastian Caldas, Sai Meher~Karthik Duddu, Peter Wu, Tian Li, Jakub
  Kone{\v{c}}n{\`y}, H~Brendan McMahan, Virginia Smith, and Ameet Talwalkar.
\newblock Leaf: A benchmark for federated settings.
\newblock {\em arXiv preprint arXiv:1812.01097}, 2018.

\bibitem{cao2021provably}
Xiaoyu Cao, Jinyuan Jia, and Neil~Zhenqiang Gong.
\newblock Provably secure federated learning against malicious clients.
\newblock In {\em Proceedings of the AAAI Conference on Artificial
  Intelligence}, volume~35, pages 6885--6893, 2021.

\bibitem{carlini2016defensive}
Nicholas Carlini and David Wagner.
\newblock Defensive distillation is not robust to adversarial examples.
\newblock {\em arXiv preprint arXiv:1607.04311}, 2016.

\bibitem{carlini2017towards}
Nicholas Carlini and David~A. Wagner.
\newblock Towards evaluating the robustness of neural networks.
\newblock In {\em 2017 {IEEE} Symposium on Security and Privacy, {SP} 2017, San
  Jose, CA, USA, May 22-26, 2017}, pages 39--57. {IEEE} Computer Society, 2017.

\bibitem{cohen2019certified}
Jeremy Cohen, Elan Rosenfeld, and Zico Kolter.
\newblock Certified adversarial robustness via randomized smoothing.
\newblock In {\em International Conference on Machine Learning}, pages
  1310--1320. PMLR, 2019.

\bibitem{gehr2018ai2}
Timon Gehr, Matthew Mirman, Dana Drachsler{-}Cohen, Petar Tsankov, Swarat
  Chaudhuri, and Martin~T. Vechev.
\newblock {AI2:} safety and robustness certification of neural networks with
  abstract interpretation.
\newblock In {\em 2018 {IEEE} Symposium on Security and Privacy, {SP} 2018,
  Proceedings, 21-23 May 2018, San Francisco, California, {USA}}, pages 3--18.
  {IEEE} Computer Society, 2018.

\bibitem{guerraoui2018hidden}
Rachid Guerraoui, S{\'e}bastien Rouault, et~al.
\newblock The hidden vulnerability of distributed learning in byzantium.
\newblock In {\em International Conference on Machine Learning}, pages
  3521--3530. PMLR, 2018.

\bibitem{hong2021federated}
Junyuan Hong, Haotao Wang, Zhangyang Wang, and Jiayu Zhou.
\newblock Federated robustness propagation: Sharing adversarial robustness in
  federated learning.
\newblock {\em arXiv preprint arXiv:2106.10196}, 2021.

\bibitem{kairouz2019advances}
Peter Kairouz, H~Brendan McMahan, Brendan Avent, Aur{\'e}lien Bellet, Mehdi
  Bennis, Arjun~Nitin Bhagoji, Kallista Bonawitz, Zachary Charles, Graham
  Cormode, Rachel Cummings, et~al.
\newblock Advances and open problems in federated learning.
\newblock {\em arXiv preprint arXiv:1912.04977}, 2019.

\bibitem{madry2017towards}
Aleksander Madry, Aleksandar Makelov, Ludwig Schmidt, Dimitris Tsipras, and
  Adrian Vladu.
\newblock Towards deep learning models resistant to adversarial attacks.
\newblock {\em arXiv preprint arXiv:1706.06083}, 2017.

\bibitem{mcmahan2017communication}
Brendan McMahan, Eider Moore, Daniel Ramage, Seth Hampson, and
  Blaise~Ag{\"{u}}era y~Arcas.
\newblock Communication-efficient learning of deep networks from decentralized
  data.
\newblock In {\em Proceedings of the 20th International Conference on
  Artificial Intelligence and Statistics, {AISTATS} 2017, 20-22 April 2017,
  Fort Lauderdale, FL, {USA}}, volume~54 of {\em Proceedings of Machine
  Learning Research}, pages 1273--1282. {PMLR}, 2017.

\bibitem{mine2006octagon}
Antoine Min{\'e}.
\newblock The octagon abstract domain.
\newblock {\em Higher-order and symbolic computation}, 19(1):31--100, 2006.

\bibitem{mirman2018differentiable}
Matthew Mirman, Timon Gehr, and Martin~T. Vechev.
\newblock Differentiable abstract interpretation for provably robust neural
  networks.
\newblock In {\em Proceedings of the 35th International Conference on Machine
  Learning, {ICML} 2018, Stockholmsm{\"{a}}ssan, Stockholm, Sweden, July 10-15,
  2018}, volume~80 of {\em Proceedings of Machine Learning Research}, pages
  3575--3583. {PMLR}, 2018.

\bibitem{shah2021adversarial}
Devansh Shah, Parijat Dube, Supriyo Chakraborty, and Ashish Verma.
\newblock Adversarial training in communication constrained federated learning.
\newblock {\em arXiv preprint arXiv:2103.01319}, 2021.

\bibitem{singh2018fast}
Gagandeep Singh, Timon Gehr, Matthew Mirman, Markus P{\"u}schel, and Martin~T
  Vechev.
\newblock Fast and effective robustness certification.
\newblock {\em NeurIPS}, 1(4):6, 2018.

\bibitem{singh2019abstract}
Gagandeep Singh, Timon Gehr, Markus P{\"u}schel, and Martin Vechev.
\newblock An abstract domain for certifying neural networks.
\newblock {\em Proceedings of the ACM on Programming Languages}, 3(POPL):1--30,
  2019.

\bibitem{tjeng2017evaluating}
Vincent Tjeng, Kai Xiao, and Russ Tedrake.
\newblock Evaluating robustness of neural networks with mixed integer
  programming.
\newblock {\em arXiv preprint arXiv:1711.07356}, 2017.

\bibitem{weber2020rab}
Maurice Weber, Xiaojun Xu, Bojan Karla{\v{s}}, Ce~Zhang, and Bo~Li.
\newblock Rab: Provable robustness against backdoor attacks.
\newblock {\em arXiv preprint arXiv:2003.08904}, 2020.

\bibitem{xie2021crfl}
Chulin Xie, Minghao Chen, Pin-Yu Chen, and Bo~Li.
\newblock Crfl: Certifiably robust federated learning against backdoor attacks.
\newblock {\em arXiv preprint arXiv:2106.08283}, 2021.

\bibitem{xie2020dba}
Chulin Xie, Keli Huang, Pin{-}Yu Chen, and Bo~Li.
\newblock {DBA:} distributed backdoor attacks against federated learning.
\newblock In {\em 8th International Conference on Learning Representations,
  {ICLR} 2020, Addis Ababa, Ethiopia, April 26-30, 2020}. OpenReview.net, 2020.

\bibitem{xie2018generalized}
Cong Xie, Oluwasanmi Koyejo, and Indranil Gupta.
\newblock Generalized byzantine-tolerant sgd.
\newblock {\em arXiv preprint arXiv:1802.10116}, 2018.

\bibitem{xie2019zeno}
Cong Xie, Sanmi Koyejo, and Indranil Gupta.
\newblock Zeno: Distributed stochastic gradient descent with suspicion-based
  fault-tolerance.
\newblock In {\em International Conference on Machine Learning}, pages
  6893--6901. PMLR, 2019.

\bibitem{yin2018byzantine}
Dong Yin, Yudong Chen, Ramchandran Kannan, and Peter Bartlett.
\newblock Byzantine-robust distributed learning: Towards optimal statistical
  rates.
\newblock In {\em International Conference on Machine Learning}, pages
  5650--5659. PMLR, 2018.

\bibitem{zizzo2020fat}
Giulio Zizzo, Ambrish Rawat, Mathieu Sinn, and Beat Buesser.
\newblock Fat: Federated adversarial training.
\newblock {\em arXiv preprint arXiv:2012.01791}, 2020.

\end{thebibliography}
\appendix

\section*{Appendix A}

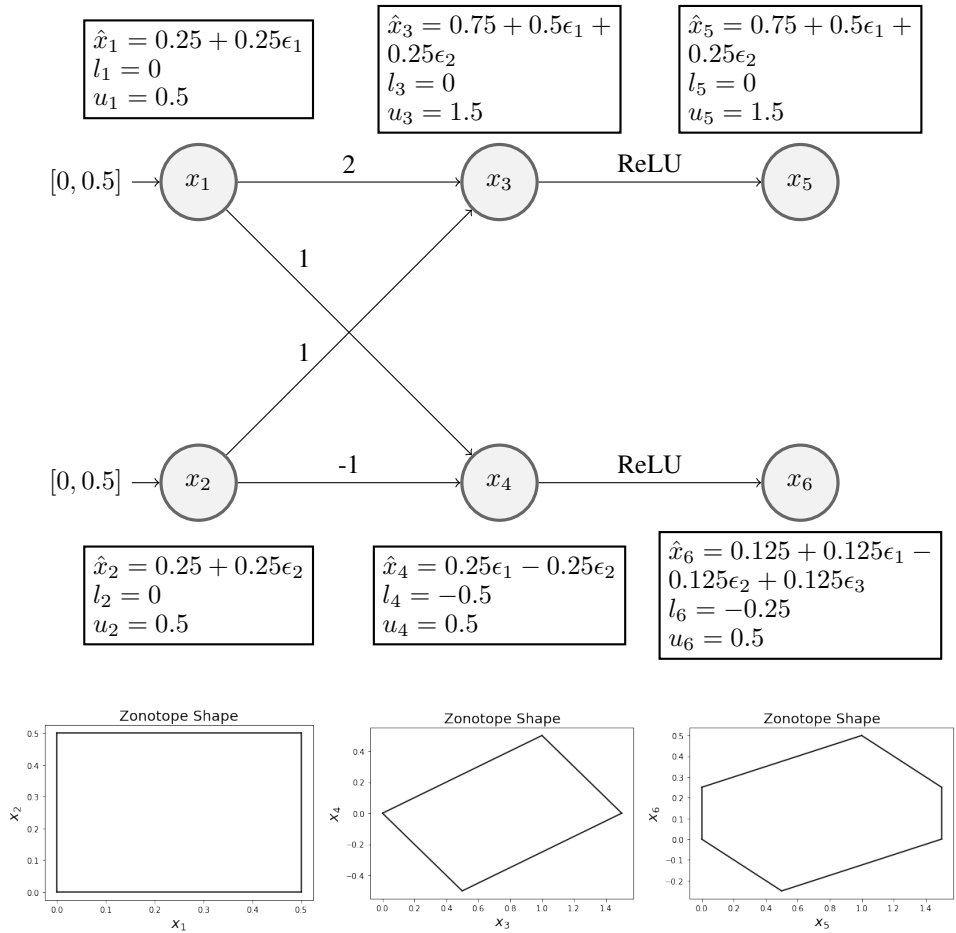
\begin{figure}
\begin{tikzpicture}[node distance=4cm,
roundnode/.style={circle, draw=black!60, fill=black!5, very thick, minimum size=10mm},]
\tikzstyle{block} = [rectangle, draw,thick,fill=blue!0,]
\node[roundnode]      (t1)                     {$x_1$};
\node[roundnode]      (b1)       [below of=t1] {$x_2$};

\node[roundnode]      (t2)       [right of=t1] {$x_3$};
\node[roundnode]      (b2)       [right of=b1] {$x_4$};

\node[roundnode]      (t3)       [right of=t2] {$x_5$};
\node[roundnode]      (b3)       [right of=b2] {$x_6$};

\node (f1)       [left of=t1, xshift=2.5cm]{$[0, 0.5]$}; 
\node (f2)       [left of=b1, xshift=2.5cm]{$[0, 0.5]$}; 

\node [block,above of=t3, yshift=-2.5cm] (start) {%
   \begin{varwidth}{10em}
      $\hat{x}_5 = 0.75 + 0.5 \epsilon_1 + 0.25 \epsilon_2 \\
      l_5 = 0 \\
      u_5 = 1.5$
    \end{varwidth}};

\node [block,above of=t2, yshift=-2.5cm] (start) {%
   \begin{varwidth}{10em}
      $\hat{x}_3 = 0.75 + 0.5 \epsilon_1 + 0.25 \epsilon_2 \\
      l_3 = 0 \\
      u_3 = 1.5$
    \end{varwidth}};

\node [block,above of=t1, yshift=-2.5cm] (start) {%
   \begin{varwidth}{10em}
      $\hat{x}_1 = 0.25 + 0.25 \epsilon_1 \\
      l_1 = 0 \\
      u_1 = 0.5$
    \end{varwidth}};

\node [block,below of=b3, yshift=+2.5cm] (start) {%
   \begin{varwidth}{10em}
      $\hat{x}_6 = 0.125 + 0.125 \epsilon_1 - 0.125 \epsilon_2 + 0.125 \epsilon_3 \\
      l_6 = -0.25 \\
      u_6 = 0.5$
    \end{varwidth}};

\node [block,below of=b2, yshift=+2.5cm] (start) {%
   \begin{varwidth}{10em}
      $\hat{x}_4 = 0.25 \epsilon_1 - 0.25 \epsilon_2 \\
      l_4 = -0.5 \\
      u_4 = 0.5$
    \end{varwidth}};

\node [block,below of=b1, yshift=+2.5cm] (start) {%
   \begin{varwidth}{10em}
      $\hat{x}_2 = 0.25 + 0.25 \epsilon_2 \\
      l_2 = 0 \\
      u_2 = 0.5$
    \end{varwidth}};

\node [below of=b1, xshift=-0.5cm, yshift=-0.5cm] {\includegraphics[width=.3\textwidth]{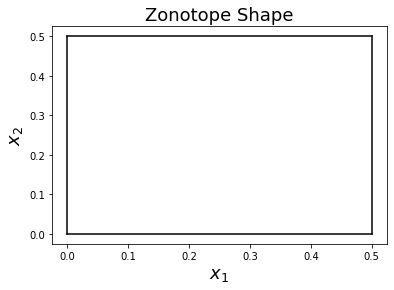}};
\node [below of=b2, xshift=-0.25cm, yshift=-0.5cm] {\includegraphics[width=.3\textwidth]{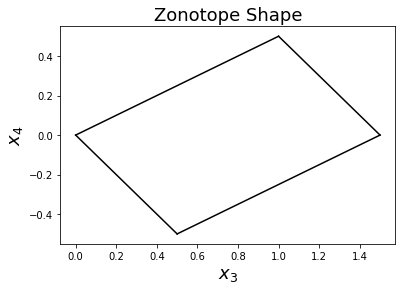}};
\node [below of=b3, xshift=0cm, yshift=-0.5cm] {\includegraphics[width=.3\textwidth]{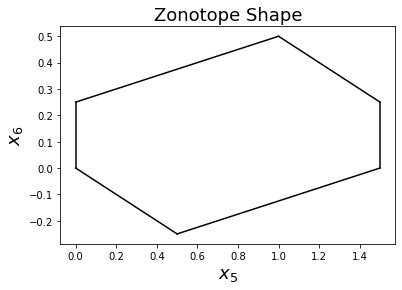}};
\draw[->] (f1.east) -- (t1.west);
\draw[->] (f2.east) -- (b1.west);

\draw[->] (t1.east)  -- node[anchor=south] {2} (t2.west);
\draw[->] (b1.east)  -- node[anchor=south] {-1} (b2.west);
\draw[->] (b1.north east) -- node[anchor=south, yshift=0.75cm, xshift=-0.6cm] {1}  (t2.south west);
\draw[->] (t1.south east) -- node[anchor=south, yshift=-0.5cm, xshift=-0.6cm] {1}  (b2.north west);

\draw[->] (t2.east)  -- node[anchor=south] {ReLU} (t3.west);
\draw[->] (b2.east)  -- node[anchor=south] {ReLU} (b3.west);

\end{tikzpicture}
\caption{Example of zonotopes being pushed through a toy layer, taking in 2 features and having a total of 4 weights. In each box we state the zonotope component as well as its upper $u$ and lower $l$ bounds.}
\label{fig:zono_example}
\end{figure}

We show in Figure \ref{fig:zono_example} an example of a zonotope being pushed through a single layer in a toy neural network to help build intuition for the reader into the ideas behind using zonotopes. In Figure \ref{fig:zono_example} we start with two features which can each take values between 0 - 0.5. We initially convert this into a zonotope, so each feature now has a central term of 0.25, and an error term $\epsilon_{1}$ or $\epsilon_{2}$ with coefficients of 0.25. With the $\epsilon$ terms being able to take $\pm1$ this draws the zonotope shape shown of a square. 

This zonotope now gets pushed through the dense layer with weights
\begin{equation}
    W = \begin{pmatrix}
        2 & 1 \\
        1 & -1
        \end{pmatrix}.
\end{equation}

Multiplication of a zonotope by a constant affects all the terms, and we add zonotopes together term by term giving us $\hat{x}_3$ and $\hat{x}_4$. This gives the rotated rectangle shown in the middle plot. Note, that here we can see a advantage of the zonotope domain over just using intervals, as the rotated rectangle has a smaller area then just considering the upper and lower bounds of $\hat{x}_3$ and $\hat{x}_4$. 

Finally, we apply a ReLU activation. As $\hat{x}_3$ has a positive upper bound and a lower bound of 0, the ReLU activation does not affect it. However, $\hat{x}_4$ has a negative lower bound and an upper bound above 0. Therefore, the application of a ReLU cannot be exactly captured- a concrete datapoint could either result in  output of 0 from the ReLU or be unaffected. To include all the options in our analysis we use the DeepZono relaxation proposed in \cite{singh2018fast} and the application of it is shown for $\hat{x}_6$ (for the details of the calculation, interested readers can find the formulation in the paper \cite{singh2018fast}). The key take away is we now introduce a new error term $\epsilon_3$ which makes the analysis more imprecise. Further, the analysis is also now more complex as we need to track more error terms and this is reflected in the final shape being composed of 3 pairs of parallel lines.

\clearpage
\section*{Appendix B}
\label{appendix_b}
MNIST and CIFAR10 images with the backdoors used for the results in Table \ref{tab:mnist_results}. We use $L_\infty$ bounds for the backdoor to follow the attacker model adversarial training is designed to protect against. In other words, if we allowed $L_0$ perturbations then backdooring to circumvent $L_\infty$ based adversarial training would not even be required as adversarial examples can be found which can evade the adversarial training. Hence, this represents perturbations which falls into what PGD should be able to protect against, and yet backdooring in this style can be used to introduce the hidden weaknesses desired by the attacker.

\begin{figure}[h!]
     \centering
     \begin{subfigure}[b]{0.45\textwidth}
         \centering
         \includegraphics[width=\textwidth]{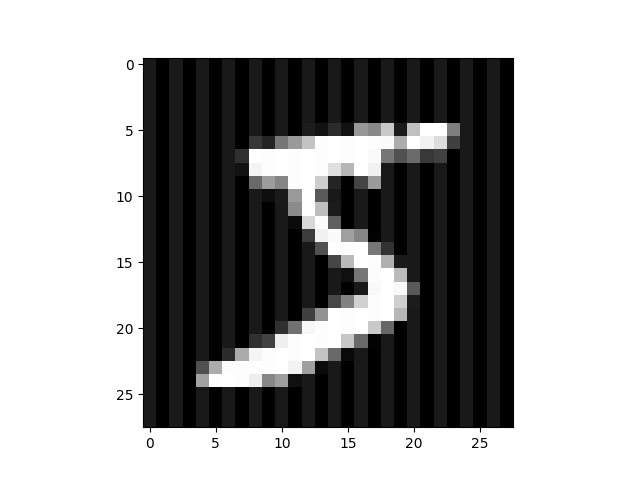}
         \caption{Vertical stripes of magnitude $L_\infty = 0.1$. Model is adversarially trained to $L_\infty = 0.25$}
         \label{fig:mnist_backdoor}
     \end{subfigure}
     \hfill
     \begin{subfigure}[b]{0.45\textwidth}
         \centering
         \includegraphics[width=\textwidth]{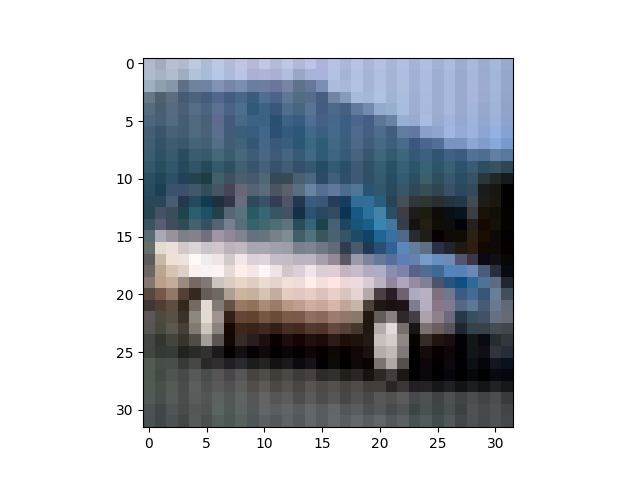}
         \caption{Vertical stripes of magnitude $L_\infty = 6 / 255$. Model is adversarially trained to $L_\infty = 8/255$}
         \label{fig:cifar_backdoor}
     \end{subfigure}
     \caption{$L_\infty$ constrained backdoor keys.}
     \label{fig:backdoors}
\end{figure}

\section*{Appendix C}

Here we have a brief overview of the functioning of defensive distillation and how it can be used as an attack. This was first proposed in the paper~\cite{zizzo2020fat}, and as it involves a few more steps compared to a backdoor style attack we detail the algorithm below.

\textbf{How do you train a network with defensive distillation?} In defensive distillation we train two neural networks: a teacher neural network and a student neural network. We apply a temperature scaling $T$ to the softmax of both networks. We first train the teacher neural network on data-label pairs $(x,y)$. After the teacher neural network has converged we use its predictions on the training dataset to re-label it. The student neural network still with temperature $T$ on the softmax is trained on the re-labeled dataset. Then, once converged, the student neural network's temperature is set back to 1. The defence is completed with the student neural network being the defended model.

\textbf{How does it protect against attack?} The core component is the temperature scaling factor which we apply during training, but remove at test time. At inference, this will case the inputs to the softmax function to be $T$ times larger. Due to this scaling, the output of the softmax will be close to a one hot vector. In fact, the components will be so close to either 0 or 1 that the 32-bit floating-point of non-predicted classes is rounded to 0 while the predicted class has a prediction of $\sim1$. For similar reasons the the gradient, when represented as 32-bit floating-point, is rounded to 0 preventing the adversarial attack from making progress \cite{carlini2017towards}.    

\textbf{How can an attacker circumvent it?:} Defensive distillation is an example of a \emph{gradient masking} defence and so has a few shortcomings. Firstly it is vulnerable against black-box transfer attacks. However, stronger attacks can be performed by an attacker who knows the details of defensive distillation. Specifically, as was noted in~\cite{carlini2016defensive} the key component of defensive distillation is the temperature scaling to cause vanishing gradients. An attacker can compute the adversarial examples by re-applying the temperature $T$ which re-scales the logits to appropriate values restoring gradient information enabling adversarial attack crafting. After crafting the adversarial examples the attacker then sends them to the original defender model with $T=1$ and the adversarial examples function just as effectively.
 
\textbf{How can it be turned into an attack in FL?} In our scenario, we can use defensive distillation to achieve the goals of the attacker set out in Section \ref{sec:attacker_model}. The model will look robust to a defender who evaluates it via standard application of the PGD attack. However, the attacker who knows that this is a defensively distilled model can craft their adversarial examples using temperature scaling as described above breaking the model.

\section*{Appendix D}

Neural network architectures used for our experiments. We use the following notation to describe our models: Conv$_{(s_w, s_h)}$ $C$ x $W$ x $H$ for a convolutional layer with $C$ output channels, and a kernel of width $W$ and height $H$. The stride size in width and height is given by  $s_w$ and $s_h$ respectively. Additionally, FC $N$ indicates a fully connected layer with $N$ outputs.

MNIST model:

Conv$_{(2,2)}$ 16x4x4 $\rightarrow$ ReLU $\rightarrow$  Conv$_{(2,2)}$ 32x4x4 $\rightarrow$  ReLU $\rightarrow$  FC 1000 $\rightarrow$  ReLU $\rightarrow$ FC 10 

CIFAR10 model: 

Conv$_{(1,1)}$ 32x3x3 $\rightarrow$ ReLU $\rightarrow$
Conv$_{(2,2)}$ 64x4x4 $\rightarrow$ ReLU $\rightarrow$
Conv$_{(1,1)}$ 64x3x3 $\rightarrow$ ReLU $\rightarrow$
Conv$_{(2,2)}$ 128x4x4 $\rightarrow$ ReLU $\rightarrow$
FC 512 $\rightarrow$ ReLU $\rightarrow$ 
FC 512 $\rightarrow$ ReLU $\rightarrow$
FC 10

\end{document}